\pdfoutput=1

\documentclass[letterpaper, 10 pt, conference]{ieeeconf}  

\IEEEoverridecommandlockouts                              

\usepackage{multibib}
\newcites{suppl}{Supplementary References}
\usepackage{multirow}
\usepackage{datetime2}
\DTMsetup{
	datesep={},
	timesep={}
}

\usepackage{graphics} 
\usepackage{amsmath} 
\usepackage{amssymb}  
\usepackage{booktabs}
\usepackage{tabularx}
\usepackage[font={footnotesize}, skip=5pt]{caption}
\usepackage{subcaption}
\usepackage{microtype}
\usepackage{flushend}
\usepackage{cuted}
\let\labelindent\relax
\usepackage[inline]{enumitem}
\renewcommand{\baselinestretch}{0.99}
\usepackage{diagbox}

\usepackage[T1]{fontenc}

\usepackage{graphicx}

\graphicspath{{figures/}}

\renewcommand{\eqref}[1]{Eq.~(\ref{#1})}

\usepackage[noadjust]{cite}
\usepackage{hyperref}
\usepackage{xcolor} 
\definecolor{mydeepblue}{rgb}{0, 0.47, 0.69}

\hypersetup{
    colorlinks=true,
    urlcolor=mydeepblue,      
}
\usepackage{pgfplots}
\pgfplotsset{compat=1.18}
\usepackage{stfloats}

\usepackage{siunitx}

\usepackage[abs,unit=1mm]{overpic}

\usepackage{csquotes}

\usepackage{float}

\newsavebox{\tempbox}

\usepackage{xspace}

\title{\LARGE \bf
DiffuDepGrasp: Diffusion-based Depth Noise Modeling Empowers Sim2Real Robotic Grasping
}

\author{Yingting Zhou$^{1,2,3}$, Wenbo Cui$^{1,2}$, Weiheng Liu$^{1,2}$, Guixing Chen$^{3}$, Haoran Li$^{1,2,3\dagger}$, and Dongbin Zhao$^{1,2}$
\thanks{$^{1}$The State Key Laboratory of Multimodal Artificial Intelligence Systems, Institute of Automation, Chinese Academy of Sciences, Beijing, China}
    \thanks{$^{2}$School of Artificial Intelligence, University of Chinese Academy of Sciences, Beijing, China}
    \thanks{$^{3}$Beijing Zhiwangweilai Technology Co., Ltd., Beijing, China}.
    \thanks{$^{\dagger}$Corresponding author.}}

\begin{document}
\maketitle
\thispagestyle{empty}
\pagestyle{empty}

\begin{abstract}
Transferring the depth-based end-to-end policy trained in simulation to physical robots can yield an efficient and robust grasping policy,
yet sensor artifacts in real depth maps like voids and noise establish a significant sim2real gap that critically impedes policy transfer. 
Training-time strategies like procedural noise injection or learned mappings suffer from data inefficiency due to unrealistic noise simulation, which is often ineffective for grasping tasks that require fine manipulation or dependency on paired datasets heavily. Furthermore, leveraging foundation models to reduce the sim2real gap via intermediate representations fails to mitigate the domain shift fully and adds computational overhead during deployment. This work confronts dual challenges of data inefficiency and deployment complexity. We propose DiffuDepGrasp, a deploy-efficient sim2real framework enabling zero-shot transfer through simulation-exclusive policy training. Its core innovation, the Diffusion Depth Generator, synthesizes geometrically pristine simulation depth with learned sensor-realistic noise via two synergistic modules. The first Diffusion Depth Module leverages temporal geometric priors to enable sample-efficient training of a conditional diffusion model that captures complex sensor noise distributions, while the second Noise Grafting Module preserves metric accuracy during perceptual artifact injection. With only raw depth inputs during deployment, DiffuDepGrasp eliminates computational overhead and achieves a $95.7$\% average success rate on $12$-object grasping with zero-shot transfer and strong generalization to unseen objects. 
Project website: \href{https://diffudepgrasp.github.io/}{\textit{https://diffudepgrasp.github.io/}}.

\end{abstract}

\section{INTRODUCTION}

Robotic grasping is a fundamental capability for achieving autonomous manipulation in unstructured real-world environments. 
A comprehensive understanding of both the geometric structure and semantic information of the scene is necessary to build a robust grasping policy.
Compared to color images, depth maps and other geometric sensor data, such as point clouds, are inherently resilient to variations in color and texture of the scene~\cite{chen2025robohanger} and provide rich and distinct geometric information.
Consequently, many imitation learning (IL) methods~\cite{Ze2024DP3, shridhar2022peract} have explored using depth or point cloud data to learn reactive policies. 
Since such spatial data are often lacking in existing real-world datasets, training in simulation and then transferring to the real world has become a cost-effective alternative to recollecting large amounts of real-world data. 
The remaining challenge lies in mitigating the discrepancies in geometric sensing data between simulation and the real world. For example, real depth maps often contain holes and noise caused by sensor imaging characteristics, which cannot be simulated in virtual environments but can significantly affect policy performance~\cite{optical, wei2024droma}.

There are roughly two ways to mitigate the simulation-to-reality (sim2real) gap faced by depth-based end-to-end policies. 
The first pathway attempts to solve the problem at training time, for instance, by injecting noise into the simulation~\cite{chen2023visual,lum2024dextrahg,dalal2024local} to enhance data realism. While injecting procedural random noise is a common strategy, it often fails to capture the complexity of real-world data. More recent data-driven approaches~\cite{xu2025sim2real} attempt to learn an implicit mapping, but the requirement for extensive computational resources and paired real-simulation data makes the overall framework costly to establish. 
An alternative pathway addresses the problem at deployment time by restoring real-world observations~\cite{zhang2024dexgraspnet} or by introducing a unified intermediate geometric representation~\cite{liu2025fetchbot}. However, these methods not only struggle to eliminate the depth domain gap completely but also introduce additional computational latency. Notably, the effectiveness of the intermediate representation strategy is highly dependent on the performance of its underlying foundation model~\cite{yang2024depth}, which has inherent limitations in metric accuracy and temporal consistency.
Consequently, building a sim2real framework that can simulate physically realistic sensor artifacts with high data efficiency and without incurring computational overhead at deployment remains an open challenge.

To address the aforementioned challenges, we propose DiffuDepGrasp, a framework for zero-shot sim2real transfer in depth-based robotic grasping.
To simulate photorealistic sensor noise while simultaneously resolving the dependency on large-scale, paired synthetic-real datasets, we design the Diffusion Depth Generator comprising two core components. 
The first Diffusion Depth Module features a conditional diffusion model that is guided by the output of a pre-trained foundation depth model~\cite{chen2025video}. This process ensures our geometric prior is both temporally stable and consistent, effectively resolving the scale ambiguity inherent in single-image estimators~\cite{yang2024depth}. Furthermore, our diffusion model learns the complex distribution of real-world sensor noise from only a small, unpaired set of RGB-D data, which significantly reduces data requirements and collection costs.
To preserve perfect geometric accuracy while simulating this noise, we design the Noise Grafting Module in the Diffusion Depth Generator, which combines the learned noise patterns with the ground-truth depth from the simulator.
The DiffuDepGrasp framework avoids additional computational latency at deployment and reduces reliance on high-performance hardware, as the policy is trained offline on the generated data and requires no extra processing of the depth input at test time. 
The simulation results and real-world results demonstrate strong zero-shot transfer and generalization capabilities.

In summary, we make the following contributions:
\begin{itemize}
    \item We propose a deploy-efficient sim2real framework, DiffuDepGrasp, which trains the policy in simulation and achieves strong zero-shot performance in the real world without additional training.

    \item We design the Diffusion Depth Generator (DDG) to generate depth maps from simulation data that approximate real-world observations with high fidelity. DDG comprises two core components: the Diffusion Depth Module exploits temporal geometric priors from minimal real RGB-D data, while the Noise Grafting Module transfers learned artifacts into simulation depth maps, preserving metric-accurate structures without compromising perceptual realism.
    
    \item The performance of our framework in the real world demonstrates that our method achieves zero-shot sim2real transfer and a $95.7$\% average success rate on the $12$-object grasping task.
\end{itemize}

\section{RELATED WORK}\label{sec:related_work}

\begin{figure*}[tb]
    \includegraphics[width=\textwidth]{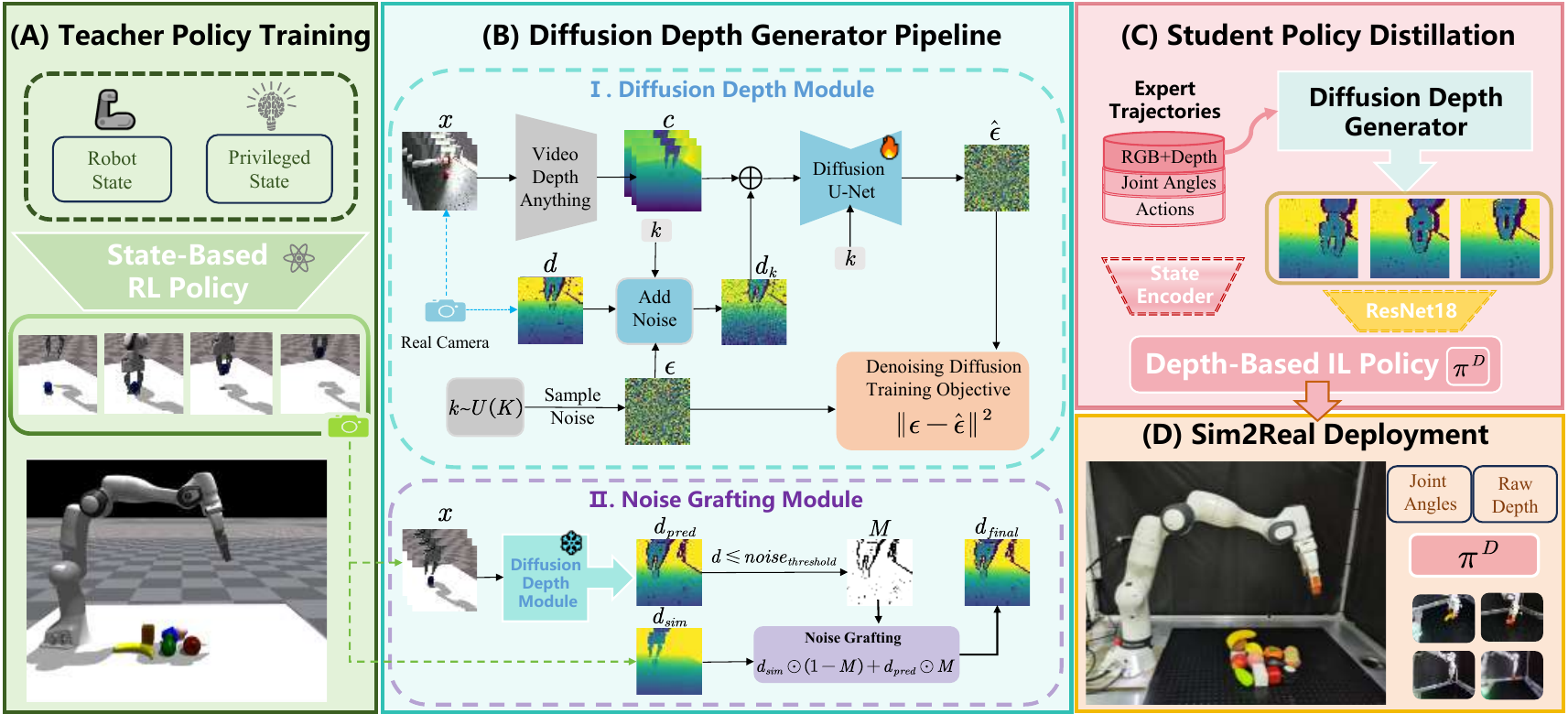}
    \caption{
    \textbf{DiffuDepGrasp Framework. }
    In the \textbf{(A) Teacher Policy Training} stage, we leverage privileged state information in simulation to train a high-performance, RL-based teacher policy for collecting expert demonstrations. 
    The \textbf{(B) Diffusion Depth Generator (DDG)} stage consists of two core modules. The first \textbf{Diffusion Depth Module} is trained on real-world, collected RGB-D data to learn the sensor's noise distribution. Note: $k$ denotes the diffusion process timestep, distinct from policy timestep $t$ in Sec.~\ref{sec:rl}. The second \textbf{Noise Grafting Module} is designed to inject these learned artifacts into pristine simulation geometry. During inference, the complete DDG algorithm transforms simulated RGB-D data into high-fidelity, noisy depth maps.
    In the \textbf{(C) Student Policy Distillation} stage, we collect expert trajectories, convert their visual data into our generated noisy depth, a process facilitated by our Diffusion Depth Generator (B), and then distill the teacher's knowledge into a student policy via imitation learning.
    Finally, this student policy achieves zero-shot \textbf{(D) Sim2Real Deployment}, transferring directly to a physical robot to perform grasping tasks. 
    }\label{fig:frame}
    \vspace{-0.3cm}
\end{figure*}

\subsection{Sim2Real Transfer for Robotic Policies}
Sim2real transfer
 is a central challenge in data-driven robotics, spanning a wide range of learning paradigms from reinforcement and imitation learning to the fine-tuning of large-scale Vision-Language-Action (VLA) models~\cite{chen2025conrft}. 
 The key to all these approaches lies in effectively overcoming both the observation gap and the physics gap between
 simulated and real-world environments~\cite{ibarz2021train,
zhao2020sim, liu2025manipulation}.
On one hand, the dynamics gap arises from mismatches between the simulated physics and real-world physical laws, such as differences in contact forces, friction, and actuator delays. A widely adopted strategy to address this is Domain Randomization (DR). 
Excellent transfer performance has been achieved by forcing the policy to become robust to physical uncertainties through training over a wide distribution of randomized dynamics parameters~\cite{xue2025unified}.
A complementary approach is System Identification~\cite{chebotar2019closing}, which aims to reduce the gap at its source by iteratively calibrating simulator parameters using a small amount of real-world data.
On the other hand, the observation gap, particularly in visual perception, is another major hurdle. For policies relying on RGB images, this gap manifests as differences in texture, lighting, and camera optics. Similar to its application in dynamics, visual domain randomization is widely used in simulation to cover real-world visual variations~\cite{sadeghi2017cadrl,tobin2017domain}. 
For policies that rely on 3D geometry (e.g., point clouds, depth maps), this gap primarily manifests as the distortion or loss of geometric information due to physical sensor limitations. A common approach to tackle this is to augment simulation data with procedural random noise to enhance policy robustness~\cite{dalal2024local}. However, such noise patterns often deviate from physical reality, leading to limited generalization in fine-grained manipulation tasks. An alternative philosophy is to make reality closer to simulation, exemplified by DexGraspNet 2.0~\cite{zhang2024dexgraspnet}, which incorporates a test-time depth restoration module to repair real depth maps. While this can improve performance, the computational latency introduced by its online restoration creates a bottleneck for tasks requiring high-frequency, real-time control.

\subsection{Bridging the Sim2Real Gap for Depth Perception}
To address the reality gap in depth perception, the research focus has gradually shifted from simple augmentation and restoration to more sophisticated techniques in data generation and representation alignment.
Early explorations centered on physics-based sensor modeling~\cite{optical,light,Meister2013SimulationOT}. These methods attempt to build explicit physical models of active light sensors within simulators to reproduce their noise characteristics. Although theoretically well-founded, they often struggle to capture the full spectrum of artifacts and their tight coupling with specific simulators limits their generalizability.
To overcome these limitations, data-driven generative methods have emerged. To overcome these limitations, data-driven generative methods have emerged, focusing on creating large-scale synthetic datasets that better capture real-world complexities. For instance, GAPartManip~\cite{11128643} tackles perception failures in articulated object manipulation through photorealistic, material-agnostic rendering. In a similar vein, but with a greater focus on the generation process itself, Stable-Sim2Real~\cite{xu2025sim2real} leverages diffusion models for photorealistic noise learning, yet its two-stage architecture and dependence on paired $3$D data limit scalability.
A different paradigm adopts a unified intermediate representation. Works like FetchBot~\cite{liu2025fetchbot} use a depth foundation model in both domains to maintain perceptual consistency. Yet, the success of this strategy is contingent on the foundation model's inherent scale ambiguity and temporal inconsistency, which can introduce new, critical issues for downstream tasks that demand precise metric information.

\section{METHOD}
This section details our sim2real robotic grasping framework (see Fig.~\ref{fig:frame}), DiffuDepGrasp, which enables zero-shot transfer by synergizing realistic data generation with an efficient policy learning pipeline.
The framework begins with training a privileged-based teacher policy via reinforcement learning in simulation (\S\ref{sec:rl}). 
Next, to bridge the critical depth gap, we propose the Diffusion Depth Generator (\S\ref{sec:Diffusion}) comprising two synergistic modules. The first Diffusion Depth Module leverages a pre-trained foundation model to ensure a temporally consistent geometric prior, and then learns to generate authentic sensor noise via a conditional diffusion model.
The second Noise Grafting Module injects the learned photorealistic artifacts onto the pristine simulation geometry to reconcile perceptual realism with geometric accuracy.
This generated data is then used to distill the teacher's knowledge into a depth-based student policy via imitation learning (\S\ref{sec:student_distillation}). Finally, the trained student policy is deployed zero-shot onto a physical robot for real-world grasping tasks (\S\ref{sec:deployment}).

\subsection{Teacher policy: reinforcement learning with privileged information}\label{sec:rl}

The learning of the teacher policy is formulated as a reinforcement learning problem where the robot observes the current observation (${o}_t$), takes an action (${a}_t$), and receives a reward ($r_t$) afterward. By leveraging the Proximal Policy Optimization (PPO)~\cite{schulman2017proximal} algorithm with massive parallelization in Isaac Gym~\cite{makoviychuk2021isaac}, the teacher policy can explore and learn an efficient and robust grasping behavior.

\subsubsection{Observation Space}
The inputs to the teacher policy, ${o}_t \in \mathbb{R}^{40}$, include the robot's proprioceptive state and task-relevant information. Specifically, it consists of joint positions ($\mathbb{R}^9$) and velocities ($\mathbb{R}^9$), the end-effector pose ($\mathbb{R}^7$), the target object pose ($\mathbb{R}^7$), the predefined target grasp pose ($\mathbb{R}^3$), and the logarithmically transformed object category ID ($\mathbb{R}^1$).

\subsubsection{Action Space}

The policy outputs an $8$D continuous action ${a}_t \in \mathbb{R}^8$. The first $7$ dimensions ${a}_{t, 1:7}$ represent the desired changes for the arm's joint positions. The final dimension ${a}_{t, 8}\in\{-1, 1\}$, is a discrete action that commands the gripper to either open ($-1$) or close ($1$). To ensure smooth and safe execution,
the final target joint position ${q}_{t+1}^{target}$ at the next time step is computed by scaling the action and adding it to the current joint position:
\begin{equation}
    {q}_{t+1}^{target} = {q}_t + s \cdot {a}_{t, 1:7}
\end{equation}
where ${q}_t$ is the current joint position and $s$ is a fixed action scale factor.

\subsubsection{Reward Design}
\label{sec:reward_function}

To maximize the success rate while ensuring the policy's behavior is both smooth and safe, we designed a densely-shaped reward function. At each timestep $t$, the complete reward function $r_t$ is defined as:
\begin{equation}
    r_t = w_1 r_{\text{reach}} + w_2 r_{\text{lift}} + w_3 r_{\text{orient}} + r_{\text{bonus}} - w_4 p_{\text{accel}}
    \label{eq:reward_total}
\end{equation}
where $w_i$ are positive weighting coefficients. Each component is designed to guide the agent through a specific stage of the grasping task.

The first term, $r_{\text{reach}}$, is a dense reward for approaching the object:
\begin{equation}
    r_{\text{reach}} = d_{t-1}^{\text{hand}} - d_{t}^{\text{hand}} 
\end{equation}
where $d_{t}^{\text{hand}}$ is a composite distance, defined as the sum of Euclidean distances between the gripper center and the object's center. This encourages the agent to pre-shape its hand for a stable grasp.

The second term, $r_{\text{lift}}$, encourages lifting the object towards a target position ${p}_{\text{target}}$:
\begin{equation}
    r_{\text{lift}} = \mathbb{I}(\text{is\_grasped}_t) \cdot (d_{t-1}^{\text{obj}} - d_{t}^{\text{obj}}) 
\end{equation}
where $d_{t}^{\text{obj}} = \|{p}_{t}^{\text{obj}} - {p}^{\text{target}}\|_2$ is the distance from the object to the target. The indicator function $\mathbb{I}(\text{is\_grasped}_t)$ is $1$ only when the gripper is closed and in close proximity to the object, ensuring that only valid lifting motions are rewarded.

The third term, $r_{\text{orient}}$, rewards the alignment of the hand's orientation with a target orientation ${q}_{\text{target}}$:
\begin{equation}
    r_{\text{orient}} = \mathbb{I}(\text{is\_lifted}_t) \cdot (\Delta\phi_{t-1} - \Delta\phi_t) 
\end{equation}
where $\Delta\phi_t$ is the angular difference (in radians) between the hand's current and target orientation. The reward is active only when the object is successfully lifted, indicated by $\mathbb{I}(\text{is\_lifted}_t)$.

A critical component is the sparse success bonus, $r_{\text{bonus}}$:
\begin{equation}
    r_{\text{bonus}} = \mathbb{I}(\text{is\_success}_t) \cdot c_1 \cdot c_2^{N_{\text{success}}} 
\end{equation}
where $\mathbb{I}(\text{is\_success}_t)$ is $1$ if the object is within a small tolerance of the target pose. $N_{\text{success}}$ is a counter for consecutive successful timesteps, and $c_1, c_2 > 1$ are constants, creating an exponentially growing reward to encourage stable goal achievement.

Finally, to promote smooth trajectories, a penalty on joint accelerations is included:
\begin{equation}
    p_{\text{accel}} = \|\ddot{{q}}_t\|^2 
\end{equation}
where $\ddot{{q}}_t$ is the acceleration of the arm joint. This term discourages jerky motions, making the learned policy safer for real-world deployment.

\subsection{Diffusion Depth Generator}
\label{sec:Diffusion}

To learn the complex noise distribution of a real-world depth sensor, we develop a method termed Diffusion Depth Generator shwon in Fig. \ref{fig:frame} (B), which generates artifact patterns characteristic of a physical sensor guided by a high-quality geometric prior (Diffusion Depth Module), and then injects the learned artifacts into
the pristine simulated geometry (Noise Grafting Module). The first module is trained on a small-scale real-world dataset comprising $80$ RGB-D trajectories of robotic interactions with $6$ distinct objects, captured by a RealSense D$455$ depth camera mounted on the robotic platform.
A key feature of our data collection process is its simplicity, as it requires neither precise object pose annotation nor alignment with CAD models.

\subsubsection{Diffusion Depth Module}
To learn real-world sensor noise artifact distributions, we design the Diffusion Depth Module. This component employs a state-of-the-art video depth estimator Video Depth Anything~\cite{chen2025video} to generate stable, temporally consistent geometric priors, circumventing flickering and scale inconsistencies inherent in single-image estimators~\cite{yang2024depth}. For each real-world RGB-D trajectory, the model processes RGB sequences into clean relative depth maps $\textit{c}$, encoding macroscopic scene geometry priors.

The backbone of our diffusion model is a noise-prediction network $\epsilon_\theta$ based on a \textbf{conditioned U-Net}~\cite{DBLP:journals/corr/RonnebergerFB15} architecture. The input of the network includes the guiding geometric prior $\textit{c}$ and the noisy depth map $\textit{d}_k$. Specifically, the U-Net employs a symmetric encoder-decoder structure with multi-level down-sampling and up-sampling blocks, interconnected by skip connections to preserve high-resolution details. Attention blocks are embedded in the middle layers to enhance the model's ability to capture long-range dependencies within the noise patterns.

During \textbf{training}, a batch of real noisy depth maps $\textit{d}_0$ and their corresponding conditional maps $\textit{c}$ are sampled from the dataset. 
A noised sample ${d}_k$ is then generated by corrupting ${d}_0$ with a random Gaussian noise $\epsilon$, where the noise level is determined by a randomly sampled timestep $k$.
The input to the U-Net is formed by concatenating $\textit{d}_k$ and $\textit{c}$ along the channel dimension, along with the time embedding for $k$. The training objective follows the standard DDPM framework of predicting the original noise $\epsilon$, with the loss function defined as:
\begin{equation}
    \mathcal{L}(\theta) = \mathbb{E}_{k, {d}_0, {c}, \epsilon} \left[ \|\epsilon - \epsilon_\theta({d}_k \oplus {c}, k)\|^2 \right]
    \label{eq:loss_conditional_ddpm}
\end{equation}
where $\oplus$ denotes concatenation. 
Through this end-to-end optimization, the diffusion model learns to denoise a random Gaussian field into a depth map that embodies complex sensor artifacts, conditioned on a clean geometric prior $\textit{c}$.

During \textbf{inference}, the process begins with a pure Gaussian noise map ${\hat{d}}_K \sim \mathcal{N}(0, I)$ and a fixed conditional depth map $c$ from the simulation. An iterative reverse denoising process is then executed for $K$ steps. At each timestep $k$ (from $K$ down to $1$), the current noisy map ${\hat{d}}_k$ is concatenated with the condition $c$ and fed into the trained network $\epsilon_\theta$ to predict the noise, $\epsilon_\theta(\hat{d}_k \oplus c, k)$. Subsequently, a sampler such as DDPM or DDIM~\cite{song2020denoising} is used to compute the less noisy map $\hat{d}_{k-1}$ from $\hat{d}_k$ based on the predicted noise. This process is repeated until $k=0$, where the final output $\hat{d}_0$ is the predicted depth map containing photorealistic noise patterns.

\subsubsection{Noise Grafting Module}
To ensure absolute geometric accuracy while simulating photorealistic noise, we designed the Noise Grafting Module. As illustrated in part II of Fig.~\ref{fig:frame} (B), the second module takes two inputs: the perceptually realistic depth map ${d}_{{pred}}$ from the Diffusion Depth Module, and the geometrically pristine simulation depth ${d}_{{sim}}$.
The module first generates a binary mask ${M}$ by identifying regions of sensor artifacts (e.g., holes) in ${d}_{{pred}}$ via a predefined threshold ${noise}_{{threshold}}$. This mask is then used to graft the artifact patterns from ${d}_{{pred}}$ onto the geometric foundation of ${d}_{{sim}}$. The final output ${d}_{{final}}$ is composed using the following formula:
\begin{equation}
    {d}_{{final}} = {d}_{{sim}} \odot ({1}-{M}) + {d}_{{pred}} \odot {M}
    \label{eq:noise_grafting}
\end{equation}
where $\odot$ denotes element-wise multiplication. This process ensures that our final generated training data retains the flawless metric geometry of the simulation while embodying the complex noise characteristics learned from the real world.

\subsection{Student policy: imitation learning from depth observations}
\label{sec:student_distillation}

To transfer the expert knowledge from the teacher to a policy that relies solely  on visual and proprioceptive information, we distill it into a student policy via imitation learning (Fig. \ref{fig:frame} (C)). 
This student policy is designed to take depth maps with realistic sensor noise as direct input, enabling zero-shot sim2real transfer.
\subsubsection{Expert Data Collection with Domain Randomization}
The distillation process begins with large-scale expert data collection in simulation.
By deploying the fully trained RL teacher policy, we record approximately $1.2$k successful grasping trajectories. Each timestep is saved as a data tuple containing the ground-truth depth map, the RGB image, the $8$-dimensional joint state, and the corresponding expert action. To ensure the policy generalizes well, we apply comprehensive domain randomization during data collection. On the dynamics side, we randomize the robot's link masses, as well as joint friction, damping, and stiffness coefficients. 
On the visual side, we apply $450\times580$ random window cropping to $1280\times720$ depth maps to enhance robustness against viewpoint shifts.
The initial joint angles and camera extrinsics (position and orientation) are also randomized within a predefined range at the start of each episode.

\subsubsection{Noisy Depth Data Generation}
To bridge the depth perception sim2real gap, we post-process the visual components of raw RGB-D datasets collected from simulation. Utilizing our pre-trained Diffusion Depth Generator, we generate corresponding high-fidelity, noisy depth maps from simulated dataset to serve as visual inputs for training depth-based student policies.

\subsubsection{Vision-State Fusion Policy Network Training}
The student policy network maps the fused visual and state information to the $8$-dimensional continuous action space. The network uses a dual-stream architecture: a vision encoder with a pre-trained ResNet-18~\cite{He2015DeepRL} backbone extracts features from the stacked depth maps, while a small MLP state encoder processes the joint angles. The feature vectors from both streams are concatenated and fed into an action head. 
Following the diffusion policy~\cite{chi2023diffusion}, 
We train the depth-based end-to-end policy, which models the conditional distribution of actions given the state. The policy network $\epsilon_\theta$ is trained to predict the noise that was added to an expert action $a_0$ at a given diffusion timestep $k$. The training objective is to minimize the Mean Squared Error (MSE) between the predicted noise $\hat{\epsilon}$ and the true Gaussian noise $\epsilon$:
\begin{equation}
    \mathcal{L}(\theta) = \mathbb{E}_{k, {s}, a_0, \epsilon} \left[ \|\epsilon - \epsilon_\theta({a}_k, k, {s})\|^2 \right]\label{eq:diffusion_policy_loss}
\end{equation}
where $s$ is the fused state representation, $a_0$ is the expert action, and ${a}_k$ is the noised action at timestep $k$. During inference, the policy generates an action by iteratively denoising from a random Gaussian vector, conditioned on the current observation.

\subsection{Zero-Shot Sim2Real Deployment}
\label{sec:deployment}

The final stage of our framework enables direct zero-shot transfer of simulation-trained student policies to physical robotic platforms (Fig. \ref{fig:frame} (D)). The deployed student policy $\pi_D$ maintains identical input specifications as during training: a temporal sequence of $3$ consecutive raw real-world depth maps coupled with $8$-dimensional joint angle states. The policy network processes these inputs in real-time through closed-loop control, generating $8$-dimensional continuous action for robotic arm grasping. The entire pipeline operates without online depth refinement, geometric estimation, or other compute-intensive preprocessing, thereby ensuring high operational efficiency and minimal hardware resource requirements.

\section{EXPERIMENTS}
In this section, we evaluate our proposed DiffuDepGrasp framework with lots of simulated and real experiments. First, we perform both qualitative and quantitative analyses to thoroughly assess the performance of the proposed Diffusion Depth Generator (DDG) and validate the realism of its generated data. Second, we compare our full framework against various sim2real baselines to demonstrate its zero-shot transfer performance and generalization capability to unseen objects. Finally, we conduct ablation studies to verify the necessity of applying the Noise Grafting Module within our framework.

\subsection{Experimental Setup and Baselines}
\label{sec:setup}
\noindent
\textbf{Platform and Environments.}
Policy training and evaluation in simulation were conducted in the Isaac Gym~\cite{makoviychuk2021isaac} physics simulator. To ensure consistency, an identical robotic hardware setup was used in both the simulation and the real world: a 7-DoF Franka Emika Panda arm equipped with a UMI parallel gripper. The simulation training set included $6$ object categories with diverse geometries and materials. To evaluate policy generalization, the test set encompassed $12$ objects divided into $2$ subsets: $6$ seen categories (consistent with training objects) and $6$ novel unseen categories with distinct morphological features. Visual perception in the real world was provided by a statically mounted Intel RealSense D$455$ depth camera at a resolution of $1280\times720$. 
During deployment, target objects were randomly placed within a \(40\,\text{cm} \times 30\,\text{cm}\) workspace on a tabletop in front of the robot. The policy operated in a closed loop at a $10$ Hz control frequency in both simulation and real-world deployment, generating continuous action commands based on depth vision and joint state feedback.

\begin{figure}[htbp]
	\centering
	\includegraphics[width=\columnwidth]{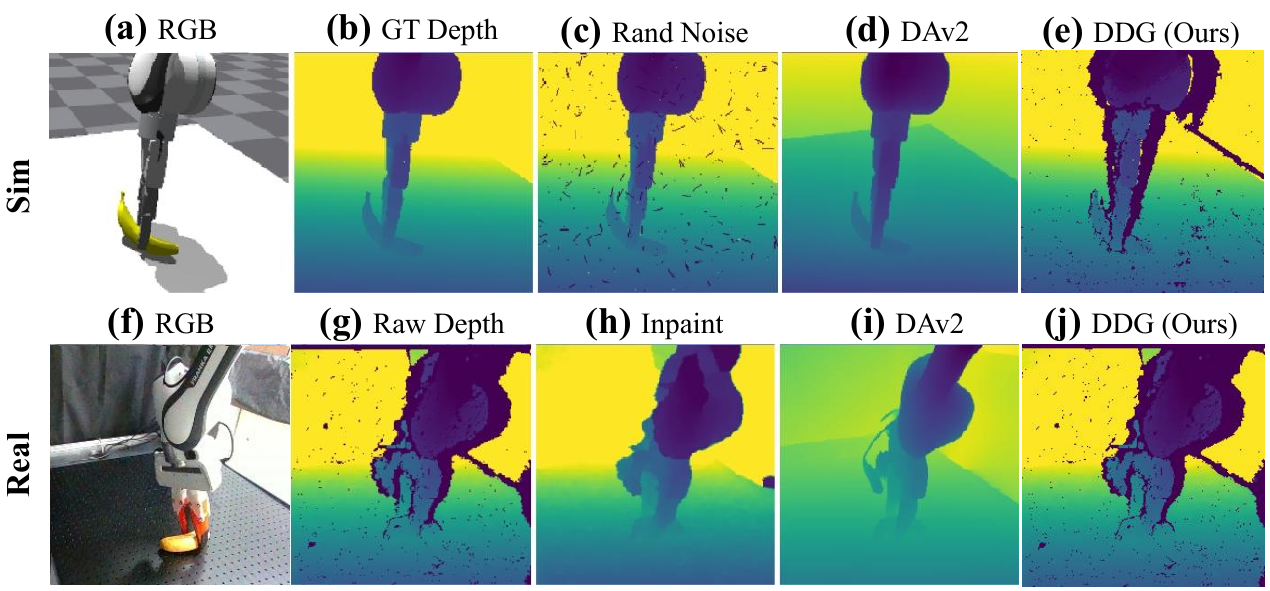} 
    \caption{\textbf{Comparison of Visual Representations for Sim2Real.}
    \textbf{(a)} Simulated RGB and \textbf{(f)} Real-world RGB.
    \textbf{(b)} Clean ground-truth (GT) depth from simulation.
    \textbf{(g)} Raw, noisy depth from the real sensor.
    The inputs  of \textbf{baselines} include: 
    \textbf{(c)} GT depth with procedural random noise (Rand Noise),
    \textbf{(h)} inpainted real depth (Inpaint),
    and \textbf{(d)},\textbf{(i)} depth estimated by DAv2 from simulated and real RGBs.
    For comparison, \textbf{(e)} and \textbf{(j)} show the final, high-fidelity depth maps generated by our proposed DDG algorithm from the simulation and real-world data, respectively.
    }\label{fig:baselines}
    \vspace{-0.3cm}
\end{figure}
   
\noindent
\textbf{Baselines}.
To evaluate the effectiveness of our proposed DiffuDepGrasp framework, we compare it against the following representative baselines (see Fig.~\ref{fig:baselines}).

\begin{itemize}
    \item \textbf{RL (State-based):} Our teacher policy, trained on privileged state information in simulation. 

\item \textbf{GT-based Policies:} This group of baselines shares the same underlying policy, trained via imitation learning on pristine, simulated ground-truth (GT) depth maps. They differ only in how they handle real-world noise:
    \begin{itemize}
        \item \textit{Naive Transfer}: The policy trained on clean data is deployed directly on raw, noisy real-world depth maps to quantify the pure domain gap.
        
        \item \textit{GT + Random Noise}~\cite{lum2024dextrahg}: Addresses the domain gap at \textit{training time} by augmenting the clean GT depth maps with procedural random noise.
        
        \item \textit{GT + Inpaint}~\cite{zhang2024dexgraspnet}: Addresses the domain gap at \textit{deployment time} by preprocessing incoming real-world depth maps with an OpenCV inpainting algorithm.
    \end{itemize}
    
\item \textbf{DAv2}~\cite{liu2025fetchbot}: The policy is trained and deployed on depth maps estimated by Depth Anything V2~\cite{yang2024depth} from simulated and real RGB images, respectively.

\end{itemize}

\subsection{Evaluation and Analysis of Depth Generation Results}

\begin{figure}[htbp]
	\centering
    \includegraphics[width=\columnwidth]{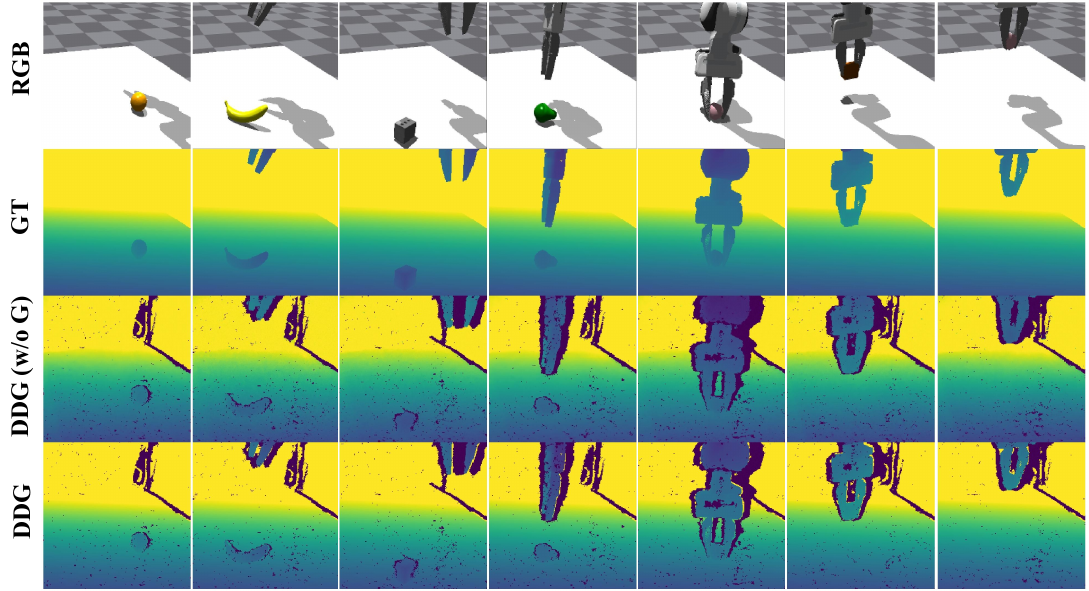} 
    \caption{
    \textbf{Qualitative Results of our Noisy Depth Data Generation.} 
    From top to bottom, the rows: 
    (1) the original simulated RGB image; 
    (2) the corresponding pristine, clean depth in simulation; 
    (3) the generated depth maps of Diffusion Depth Module without Noise Grafting Module (DDG w/o G); 
    and (4) the generated depth maps of Diffusion Depth Module with Noise Grafting Module (DDG).  }\label{fig:qualitative_result}
    \vspace{-0.3cm}
\end{figure}

We evaluate the realism of the data generated by Diffusion Depth Generator from $3$ complementary perspectives: qualitative visualization for assessing perceptual fidelity, t-SNE analysis for examining feature-space distribution alignment, and quantitative FID/KID metrics for measuring distributional divergence.
\subsubsection{Visualization of predicted depth maps} As shown in Figure~\ref{fig:qualitative_result}, DDG (w/o G) (Row 3) and DDG (Row 4) successfully generate complex and photorealistic sensor artifacts when compared to the pristine ground-truth depth from the simulator (Row 2). They not only accurately reproduce the spatial distribution and shape of holes but also render subtle noise textures along object and gripper edges. This result is in stark contrast to the uniform patterns of procedural random noise (see Fig.~\ref{fig:baselines}(c)). The
necessity of Noise Grafting Module will be further justified by the downstream task performance in our subsequent ablation study (see \S\ref{sec:ablation_grafting}).

\subsubsection{t-SNE Visualization}
To intuitively visualize the effectiveness of different methods in bridging the domain gap, we use t-SNE~\cite{JMLR:v9:vandermaaten08a} for dimensionality reduction and visualization of the depth map feature space. Specifically, we first employ a pre-trained ResNet-18~\cite{He2015DeepRL} network as a feature extractor. For $8,000$ depth samples from various domains, we extract the $512$-dimensional feature vectors from the penultimate layer (before the final classification layer) of the ResNet-18. These high-dimensional features are then non-linearly projected into a $2$D latent space using t-SNE for comparison, revealing the intricate associations within the data.

\begin{figure}[ht]
    \centering
    \includegraphics[width=\columnwidth]
    {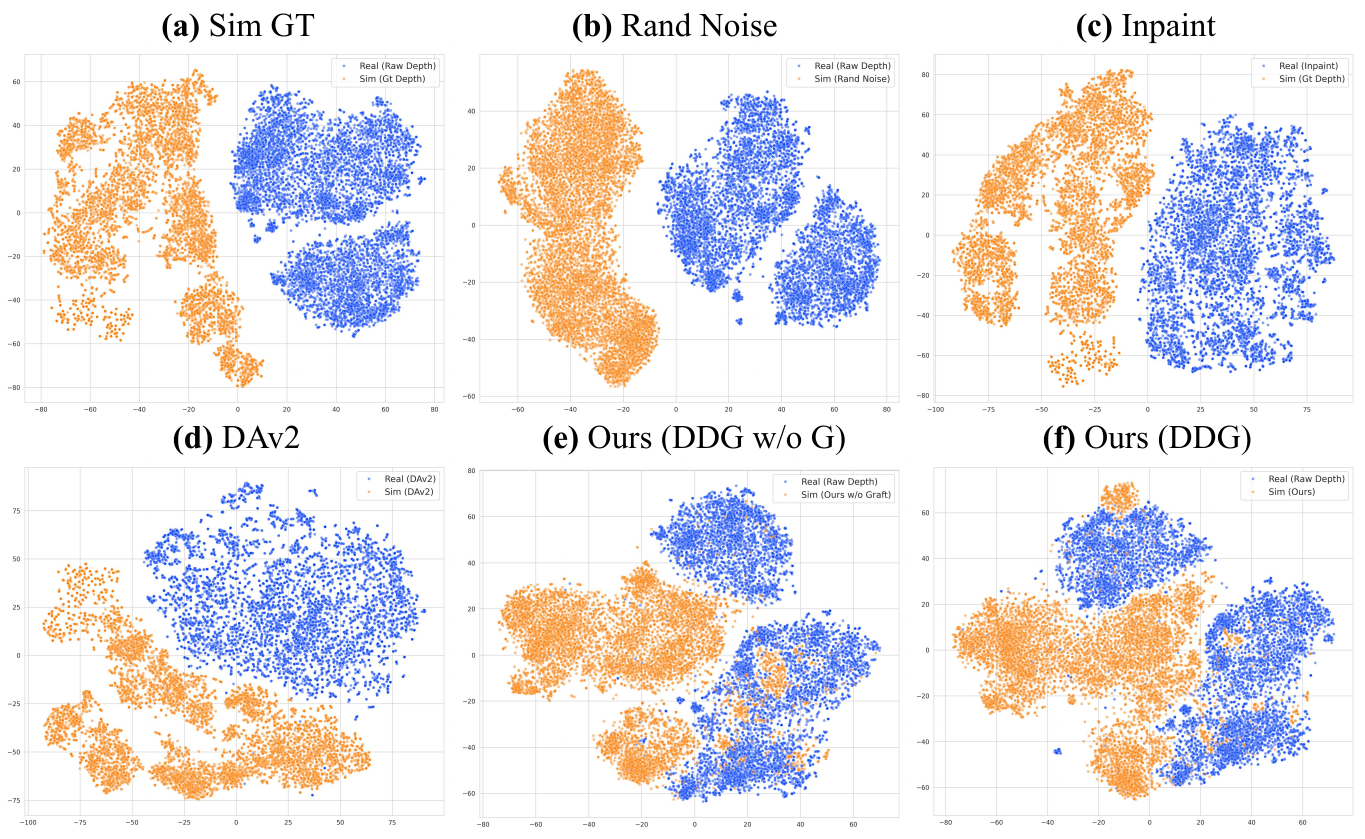} 
    \caption{
    \textbf{t-SNE Visualization.}
    Each subplot visualizes the feature distribution of  
    real-world depth data (\textcolor{blue}{blue points}) against the generated depth data via the simulation-based methods (\textcolor{orange}{orange points}).
    We define the terms as follows: 
    Real Raw: Raw depth from the physical sensor. 
    Sim GT: Clean ground-truth depth from simulation. 
    Sim Rand Noise: Sim GT depth with procedural noise.
    Real Inpaint: Real Raw depth after applying inpainting algorithm.
    Sim/Real DAv2: Depth estimated by Depth Anything V2 from sim/real RGB.
    Sim DDG w/o G: Depth generated by Diffusion Depth Module without Noise Grafting.
    Sim DDG: Depth generated by Diffusion Depth Module with Noise Grafting.
    The specific pairs compared are:
    \textbf{(a)} Real Raw vs. Sim GT;
    \textbf{(b)} Real Raw vs. Sim Random Noise;
    \textbf{(c)} Real Inpaint vs. Sim GT;
    \textbf{(d)} Real DAv2 vs. Sim DAv2;
    \textbf{(e)} Real Raw vs. Sim DDG w/o Noise Grafting;
    \textbf{(f)} Real Raw vs. Sim DDG.
}
\label{fig:t-SNE}
    \vspace{-0.3cm}
\end{figure}

\noindent
\textbf{Results and Analysis.} 
As shown in Figure~\ref{fig:t-SNE}, the simulated and real depth data (a) form distinct, well-separated clusters, visually confirming the large initial domain gap. The procedural noise method (b) remains significantly deviated from the real distribution. While the inpainting method (c) and intermediate representation-based methods like DAv2 (d) reduce the distributional distance, their boundaries remain clearly distinguishable.
In stark contrast, the direct output of our diffusion model, Ours (DDG w/o G) (e), already exhibits a significant overlap and intermingling with the real data distribution, which first validates the strong capability of our diffusion model in learning authentic noise patterns. More importantly, our final method, Ours (DDG) (f), which applies Noise Grafting, shows a similarly high degree of fusion with the real data. This key observation indicates that our Noise Grafting Module does not corrupt the photorealistic noise distribution learned by the diffusion model while ensuring geometric accuracy. Although both variants perform exceptionally well in the feature space, we ultimately choose the version with Noise Grafting, the necessity of which will be further justified in our subsequent ablation study (see \S\ref{sec:ablation_grafting}).

\begin{table}[htbp]
\centering
\caption{
    \textbf{Quantitative Comparison of Sim2Real Data Generation Methods.}
    The Sim GT, Rand Noise, and Ours columns are compared against raw real-world sensor data.
    Inpaint (Sim GT vs. Real Inpaint) and DAv2 (Sim DAv2 vs. Real DAv2) are included to evaluate the intra-domain self-consistency of their respective methods.
    KID values are scaled by $100$ for readability. Best results are in \textbf{bold}.
}
\label{tab:quantitative_comparison_real}
\resizebox{\linewidth}{!}
{
\begin{tabular}{lccccc}
\toprule
\textbf{Metric} & Sim GT & Rand Noise & Inpaint & DAv2 & \textbf{Ours} \\
\midrule
\textbf{FID ↓} & 422.24 & 242.15 & 241.67 & 109.36 & \textbf{87.01} \\
\textbf{KID ↓} & 58.45 & 31.02 & 24.33 & 8.03 & \textbf{7.09} \\
\bottomrule
\end{tabular}
}

\end{table}

\begin{table*}[ht]
\centering
\caption{
    \textbf{The Performance of Different Methods on Seen and Unseen Objects.}
    The table compares success rates (\%) in both simulation (S) and the real world (R). 
    The Seen category includes $6$ object types used during policy training; 
    The Unseen category includes $6$ novel object types used only for evaluating generalization in the real world.
    The Overall column denotes the mean success rate across $12$ objects in real-world experiments.
    N/A indicates the policy is not applicable in a given domain. Best real-world results are in \textbf{bold}.}
\label{tab:main_results_seen_unseen}
\resizebox{\textwidth}{!}{%
    \small
    \setlength{\tabcolsep}{3pt} 
    \begin{tabular}{l|cc|cc|cc|cc|cc|cc||cc||cccccc|c||c}
    \toprule
    & \multicolumn{14}{c||}{\textbf{Seen Objects}} & \multicolumn{7}{c||}{\textbf{Unseen Objects}}
    & \multicolumn{1}{c}{\textbf{Overall}} \\
    \cmidrule(r){2-15} \cmidrule(l){16-22}\cmidrule(r){23-23}

    & \multicolumn{2}{c|}{Apple} & \multicolumn{2}{c|}{Pear} & \multicolumn{2}{c|}{Peach} & \multicolumn{2}{c|}{Banana} & \multicolumn{2}{c|}{Cube} & \multicolumn{2}{c||}{Block} & \multicolumn{2}{c||}{\textbf{Avg}} & Persimmon & Starfruit & Lemon & Can & Wheel & Bread & \textbf{Avg} & \textbf{Avg} \\

    \textbf{Method} & S & R & S & R & S & R & S & R & S & R & S & R & S & R & R & R & R & R & R & R & R & R \\
    \midrule

    RL & 98.6 & N/A & 91.1 & N/A & 99.6 & N/A & 91.7 & N/A & 95.8 & N/A & 95.6 & N/A & 95.4 & N/A & \multicolumn{7}{c||}{N/A} & N/A \\
    \hline
    GT & 93.7 & 0 & 89.2 & 0 & 97.2 & 0 & 78.7 & 0 & 94.4 & 0 & 94.6 & 0 & 91.3 & 0 & 0 & 0 & 0 & 0 & 0 & 0 & 0 & 0 \\
    Rand Noise~\cite{lum2024dextrahg} & 90.8 & 0 & 90.4 & 0 & 89.6 & 0 & 68.2 & 0 & 87.0 & 0 & 87.7 & 0 & 85.6  & 0 & 0 & 0 & 0 & 0 & 0 & 0 & 0 & 0\\
    Inpaint~\cite{zhang2024dexgraspnet}  & 93.7 & 17.5 & 89.2 & 45.0 & 97.2 & 60.0 & 78.7 & 90.0 & 94.4 & 20.0 & 94.6 & 45.0 & 91.3 & 46.3 & 55.0 & 30.0 & 20.0 & 65.0 & 45.0 & 65.0 & 46.7 & 46.5\\
    DAv2~\cite{liu2025fetchbot} & 87.3 & 80.0 & 84.1 & 77.5 & 86.1 & 50.0 & 73.5 & 81.7 & 86.6 & 90.0 & 85.4 & 95.0 & 83.8 & 79.0 & 70.0 & 70.0 & 50.0 & 90.0 & 90.0 & 100.0 & 78.3 & 78.7\\

    Ours (w/o G) & N/A & 50.0 & N/A & 60.0 & N/A & 65.0 & N/A & 60.0 & N/A & 35.7 & N/A & 80.0 & 
    N/A &
    58.5 & 70.0 & 70.0 & 20.0 & 60.0 & 60.0 & 65.0 &  57.5 & 58.0\\
    
    \textbf{Ours} & N/A & \textbf{95.0} & N/A & \textbf{100.0} & N/A & \textbf{100.0} & N/A & \textbf{90.0} & N/A & \textbf{100.0} & N/A & \textbf{100.0} & N/A & \textbf{97.5} &
    \textbf{100.0} &
    \textbf{97.5} & \textbf{65.0} & \textbf{100.0} & \textbf{100.0} & \textbf{100.0} & \textbf{93.8}
    & \textbf{95.7}\\

    \bottomrule
    \end{tabular}%
} 
\end{table*}

\subsubsection{Quantitative Evaluation of Generation Quality}
To quantitatively assess the realism of the generated depth maps, we employ two widely-used distributional metrics: the Fréchet Inception Distance (FID)~\cite{Heusel2017GANsTB
} and the Kernel Inception Distance (KID)~\cite{bińkowski2018demystifying}. Both metrics compute the distance between the feature distributions of generated and real data, extracted from a pre-trained Inception network. While FID primarily compares the first two moments (mean and covariance) under a Gaussian assumption, KID provides a non-parametric alternative that is more robust to smaller sample sizes and sensitive to higher-order distribution moments. 
Lower FID and KID scores indicate higher similarity between the distributions, suggesting that the generated images are closer to real data in terms of visual quality and diversity.

\noindent
\textbf{Results and Analysis.}
As presented in Table~\ref{tab:quantitative_comparison_real}, both the ground truth simulated depth (Sim GT) and the random noise augmented data (Rand Noise) exhibit a significant distributional difference from the real sensor data. More advanced baselines like Inpaint (FID $241.67$) and DAv2 (FID $109.36$) substantially reduce this gap, but a noticeable discrepancy remains.
In contrast, our proposed DDG method achieves the best performance, obtaining the lowest FID ($87.01$) and KID ($7.09$) scores. This result provides strong quantitative evidence that our DiffuDepGrasp framework, through its diffusion-based learning and Noise Grafting Module, is superior in generating high-fidelity simulated data that most closely matches the statistical distribution of a real-world sensor.

\subsection{Evaluation of Zero-Shot Sim2Real Performance} 
\begin{figure}[htbp]
\centering
\includegraphics[width=\columnwidth]
{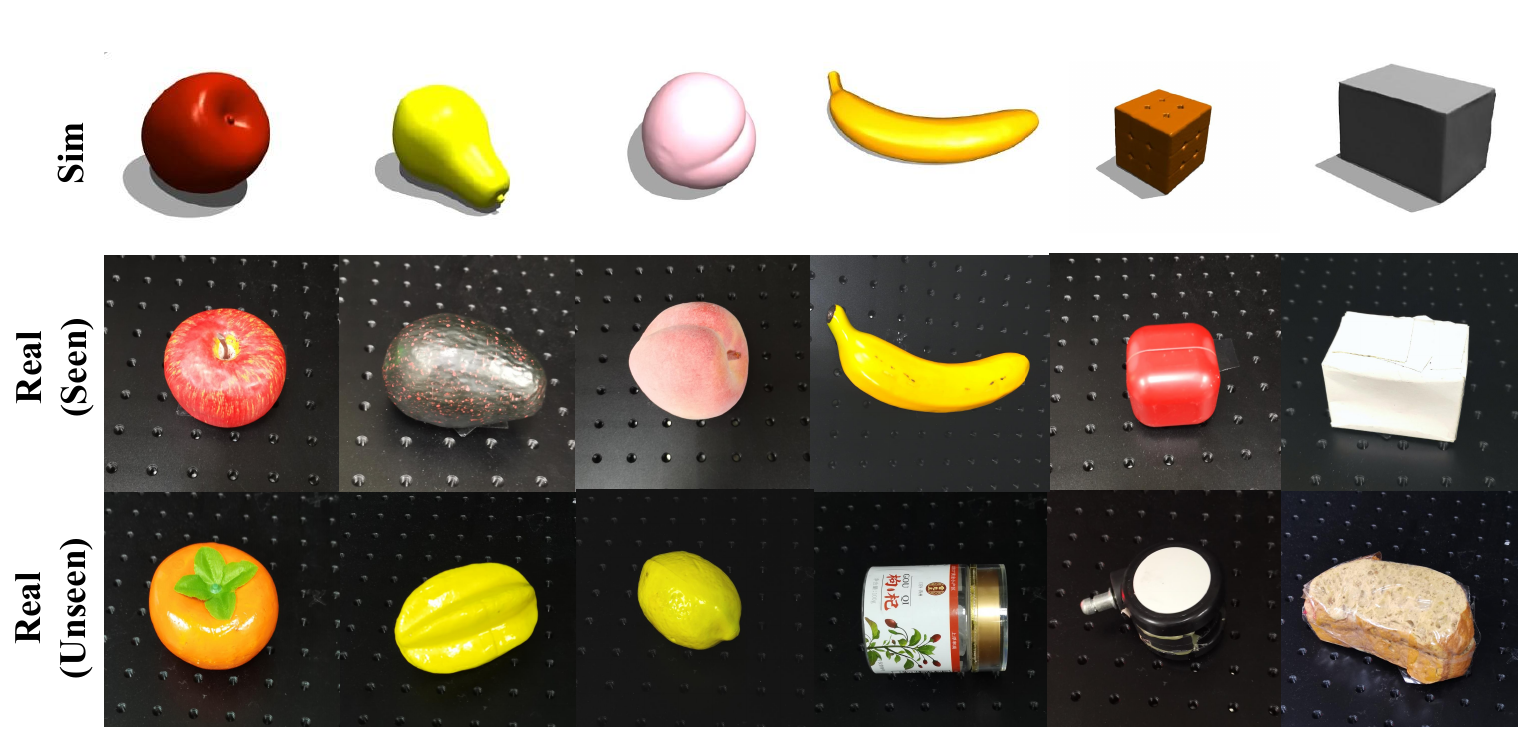} 
\caption{\textbf{The objects used to train policies in simulation and real world.}}
\label{fig:experimental_objects}
\vspace{-0.3cm}
\end{figure}

We ultimately evaluate the end-to-end performance of our framework through extensive zero-shot transfer experiments on a physical Franka robot. Figure~\ref{fig:experimental_objects} shows a subset of the objects used in our simulation training (Sim) and real-world evaluation (Real). The real-world test set is divided into $2$ groups: Seen ($6$ categories present in simulation training) and Unseen ($6$ novel categories), to assess both the fundamental performance and the generalization capability of the policies. 
Our primary evaluation metric is the \textbf{Grasping Success Rate}. A grasp attempt is considered successful if the robot securely grasps the target object and lifts it to a predefined height ($40$ cm) without dropping it. Each object category was tested for $100$ trials in simulation and $20$ trials in the real world.

\noindent
\textbf{Results and Analysis.}
Table~\ref{tab:main_results_seen_unseen} summarizes the grasping success rates of all methods. From the Seen Objects category, the severity of the sim2real gap is clearly observable. The GT and Rand Noise policies, which performed reasonably in simulation, fail in the real world ($0$\% success rate). More advanced baselines like Inpaint and DAv2 achieve some success but still suffer from a significant performance degradation compared to their simulation performance (e.g., Inpaint drops from $91.3$\% to $46.3$\%). In contrast, 
our proposed method not only avoids this performance drop but achieves a superior real-world success rate of $97.5\%$, which is higher than the simulation performance of the state-based RL teacher ($95.4\%$). We attribute this to the difference in evaluation protocols. The simulation evaluation was conducted over a much wider and more challenging distribution of initial states, whereas the real-world evaluation covered a comparatively simpler set of scenarios.
For the Unseen Objects category, our method (Ours) also maintains the highest average success rate among all approaches, despite a marginal performance drop compared to seen objects. This discrepancy primarily stems from inherent grasp challenges posed by specific novel objects (e.g., lemons with smooth, rollable surfaces). Crucially, the sustained high success rate on unseen objects validates superior generalization capability. These results demonstrate that our framework enables the policy to learn universal grasp-affording geometric features via physically plausible noise modeling, rather than overfitting to the visual appearances of training objects.

\subsection{Ablation Study of Noise Grafting}
\label{sec:ablation_grafting}

We perform a real-world ablation study to evaluate how our proposed Noise Grafting Module contributes to successful sim2real transfer and overall grasping performance. As shown in Table~\ref{tab:main_results_seen_unseen}, we compare our full method (Ours) against a variant without Noise Grafting (Ours (w/o G)). This variant trains the student policy directly on the depth maps predicted by Diffusion Depth Module, which may contain geometric inaccuracies. On seen objects, removing Noise Grafting causes the average success rate to drop sharply from $97.5$\% to $58.5$\%. On unseen objects, the success rate plummets from $93.8$\% to $57.5$\%. The significant degradation confirms that preserving metric geometry accuracy is crucial for bridging the sim2real gap. While our diffusion model can generate perceptually realistic noise, its direct geometric output is insufficient for precision manipulation tasks.

\section{Conclusion and Future Work}
\noindent
\textbf{Conclusion.}
In this paper, we present DiffuDepGrasp, a novel and deployment-efficient framework bridging the sim2real gap for depth-driven robotic grasping through a teacher-student distillation pipeline. Its core innovation, the Diffusion Depth Generator, synthesizes realistic sensor noise with geometrically accurate simulations via two synergistic modules, tackling issues of data inefficiency and deployment complexity. Real-world experiments show that DiffuDepGrasp achieves a $95.7$\% success rate in complex $12$-object grasping tasks encompassing both seen and unseen categories, significantly outperforming prior methods.

\noindent
\textbf{Future Work.}
This study identifies future research directions for generative sim2real depth modeling. While our current noise generator primarily addresses material-induced depth voids, it could be extended to model more complex sensor artifacts, such as metrical distortion caused by transparent or translucent object boundaries. Training on a diverse real-world dataset encompassing such challenging materials would enable generalization to manipulation tasks involving complex photometric interactions.

\addtolength{\textheight}{-12cm}   



\typeout{}
\footnotesize


\begin{thebibliography}{10}
\providecommand{\url}[1]{#1}
\csname url@samestyle\endcsname
\providecommand{\newblock}{\relax}
\providecommand{\bibinfo}[2]{#2}
\providecommand{\BIBentrySTDinterwordspacing}{\spaceskip=0pt\relax}
\providecommand{\BIBentryALTinterwordstretchfactor}{4}
\providecommand{\BIBentryALTinterwordspacing}{\spaceskip=\fontdimen2\font plus
\BIBentryALTinterwordstretchfactor\fontdimen3\font minus \fontdimen4\font\relax}
\providecommand{\BIBforeignlanguage}[2]{{%
\expandafter\ifx\csname l@#1\endcsname\relax
\typeout{** WARNING: IEEEtran.bst: No hyphenation pattern has been}%
\typeout{** loaded for the language `#1'. Using the pattern for}%
\typeout{** the default language instead.}%
\else
\language=\csname l@#1\endcsname
\fi
#2}}
\providecommand{\BIBdecl}{\relax}
\BIBdecl

\bibitem{chen2025robohanger}
Y.~Chen, S.~Wei, B.~Xiao, J.~Lyu, J.~Chen, F.~Zhu, and H.~Wang, ``Robohanger: Learning generalizable robotic hanger insertion for diverse garments,'' \emph{IEEE Robotics and Automation Letters (RAL)}, 2025.

\bibitem{Ze2024DP3}
Y.~Ze, G.~Zhang, K.~Zhang, C.~Hu, M.~Wang, and H.~Xu, ``3d diffusion policy: Generalizable visuomotor policy learning via simple 3d representations,'' in \emph{Proceedings of Robotics: Science and Systems (RSS)}, 2024.

\bibitem{shridhar2022peract}
M.~Shridhar, L.~Manuelli, and D.~Fox, ``Perceiver-actor: A multi-task transformer for robotic manipulation,'' in \emph{Proceedings of the 6th Conference on Robot Learning (CoRL)}, 2022.

\bibitem{optical}
X.~Zhang, R.~Chen, A.~Li, F.~Xiang, Y.~Qin, J.~Gu, Z.~Ling, M.~Liu, P.~Zeng, S.~Han, Z.~Huang, T.~Mu, J.~Xu, and H.~Su, ``Close the optical sensing domain gap by physics-grounded active stereo sensor simulation,'' \emph{IEEE Transactions on Robotics}, 2023.

\bibitem{wei2024droma}
S.~Wei, H.~Geng, J.~Chen, C.~Deng, C.~Wenbo, C.~Zhao, X.~Fang, L.~Guibas, and H.~Wang, ``D3roma: Disparity diffusion-based depth sensing for material-agnostic robotic manipulation,'' in \emph{8th Annual Conference on Robot Learning}, 2024.

\bibitem{chen2023visual}
T.~Chen, M.~Tippur, S.~Wu, V.~Kumar, E.~Adelson, and P.~Agrawal, ``Visual dexterity: In-hand reorientation of novel and complex object shapes,'' \emph{Science Robotics}, vol.~8, no.~84, p. eadc9244, 2023.

\bibitem{lum2024dextrahg}
T.~G.~W. Lum, M.~Matak, V.~Makoviychuk, A.~Handa, A.~Allshire, T.~Hermans, N.~D. Ratliff, and K.~V. Wyk, ``Dextr{AH}-g: Pixels-to-action dexterous arm-hand grasping with geometric fabrics,'' in \emph{8th Annual Conference on Robot Learning (CoRL)}, 2024.

\bibitem{dalal2024local}
M.~Dalal, M.~Liu, W.~Talbott, C.~Chen, D.~Pathak, J.~Zhang, and R.~Salakhutdinov, ``Local policies enable zero-shot long-horizon manipulation,'' in \emph{CoRL Workshop on Learning Robot Fine and Dexterous Manipulation: Perception and Control}, 2024.

\bibitem{xu2025sim2real}
M.~Xu, C.~Ye, H.~Liu, Y.~Wu, J.~Chang, and X.~Han, ``Stable-sim2real: Exploring simulation of real-captured 3d data with two-stage depth diffusion,'' in \emph{International Conference on Computer Vision (ICCV)}, 2025.

\bibitem{zhang2024dexgraspnet}
J.~Zhang, H.~Liu, D.~Li, X.~Yu, H.~Geng, Y.~Ding, J.~Chen, and H.~Wang, ``Dexgraspnet 2.0: Learning generative dexterous grasping in large-scale synthetic cluttered scenes,'' in \emph{8th Annual Conference on Robot Learning (CoRL)}, 2024.

\bibitem{liu2025fetchbot}
W.~Liu, Y.~Wan, J.~Wang, Y.~Kuang, W.~Cui, X.~Shi, H.~Li, D.~Zhao, Z.~Zhang, and H.~Wang, ``Fetchbot: Learning generalizable object fetching in cluttered scenes via zero-shot sim2real,'' \emph{Conference on Robot Learning (CoRL)}, 2025.

\bibitem{yang2024depth}
L.~Yang, B.~Kang, Z.~Huang, Z.~Zhao, X.~Xu, J.~Feng, and H.~Zhao, ``Depth anything v2,'' \emph{Advances in Neural Information Processing Systems}, vol.~37, pp. 21\,875--21\,911, 2024.

\bibitem{chen2025video}
S.~Chen, H.~Guo, S.~Zhu, F.~Zhang, Z.~Huang, J.~Feng, and B.~Kang, ``Video depth anything: Consistent depth estimation for super-long videos,'' in \emph{Proceedings of the Computer Vision and Pattern Recognition Conference (CVPR)}, 2025, pp. 22\,831--22\,840.

\bibitem{chen2025conrft}
Y.~Chen, S.~Tian, Y.~Zhou, S.~Liu, H.~Li, and D.~Zhao, ``Conrft: A reinforced fine-tuning method for vla models via consistency policy,'' in \emph{Robotics: Science and Systems (RSS)}, 2025.

\bibitem{ibarz2021train}
J.~Ibarz, J.~Tan, C.~Finn, M.~Kalakrishnan, P.~Pastor, and S.~Levine, ``How to train your robot with deep reinforcement learning: lessons we have learned,'' \emph{The International Journal of Robotics Research}, vol.~40, no. 4-5, pp. 698--721, 2021.

\bibitem{zhao2020sim}
W.~Zhao, J.~P. Queralta, and T.~Westerlund, ``Sim-to-real transfer in deep reinforcement learning for robotics: a survey,'' in \emph{2020 IEEE symposium series on computational intelligence (SSCI)}.\hskip 1em plus 0.5em minus 0.4em\relax IEEE, 2020, pp. 737--744.

\bibitem{liu2025manipulation}
M.~Liu, Z.~Zhu, X.~Han, P.~Hu, H.~Lin, X.~Li, J.~Chen, J.~Xu, Y.~Yang, Y.~Lin, X.~Li, Y.~Yu, W.~Zhang, T.~Kong, and B.~Kang, ``Manipulation as in simulation: Enabling accurate geometry perception in robots,'' \emph{arXiv preprint}, 2025.

\bibitem{xue2025unified}
Y.~Xue, W.~Dong, M.~Liu, W.~Zhang, and J.~Pang, ``A unified and general humanoid whole-body controller for fine-grained locomotion,'' in \emph{Robotics: Science and Systems (RSS)}, 2025.

\bibitem{chebotar2019closing}
Y.~Chebotar, A.~Handa, V.~Makoviychuk, M.~Macklin, J.~Issac, N.~Ratliff, and D.~Fox, ``Closing the sim-to-real loop: Adapting simulation randomization with real world experience,'' in \emph{2019 International Conference on Robotics and Automation (ICRA)}.\hskip 1em plus 0.5em minus 0.4em\relax IEEE, 2019, pp. 8973--8979.

\bibitem{sadeghi2017cadrl}
F.~Sadeghi and S.~Levine, ``{CAD2RL}: Real single-image flight without a single real image,'' in \emph{Robotics: Science and Systems(RSS)}, 2017.

\bibitem{tobin2017domain}
J.~Tobin, R.~Fong, A.~Ray, J.~Schneider, W.~Zaremba, and P.~Abbeel, ``Domain randomization for transferring deep neural networks from simulation to the real world,'' in \emph{2017 IEEE/RSJ international conference on intelligent robots and systems (IROS)}.\hskip 1em plus 0.5em minus 0.4em\relax IEEE, 2017, pp. 23--30.

\bibitem{light}
K.~Bai, L.~Zhang, Z.~Chen, F.~Wan, and J.~Zhang, ``Close the sim2real gap via physically-based structured light synthetic data simulation,'' in \emph{ICRA}, 2024.

\bibitem{Meister2013SimulationOT}
S.~Meister, R.~Nair, and D.~Kondermann, ``Simulation of time-of-flight sensors using global illumination,'' in \emph{International Symposium on Vision, Modeling, and Visualization}, 2013.

\bibitem{11128643}
W.~Cui, C.~Zhao, S.~Wei, J.~Zhang, H.~Geng, Y.~Chen, H.~Li, and H.~Wang, ``Gapartmanip: A large-scale part-centric dataset for material-agnostic articulated object manipulation,'' in \emph{2025 IEEE International Conference on Robotics and Automation (ICRA)}, 2025, pp. 14\,791--14\,798.

\bibitem{schulman2017proximal}
J.~Schulman, F.~Wolski, P.~Dhariwal, A.~Radford, and O.~Klimov, ``Proximal policy optimization algorithms,'' \emph{arXiv preprint arXiv:1707.06347}, 2017.

\bibitem{makoviychuk2021isaac}
V.~Makoviychuk, L.~Wawrzyniak, Y.~Guo, M.~Lu, K.~Storey, M.~Macklin, D.~Hoeller, N.~Rudin, A.~Allshire, A.~Handa, and G.~State, ``Isaac gym: High performance {GPU} based physics simulation for robot learning,'' in \emph{Thirty-fifth Conference on Neural Information Processing Systems Datasets and Benchmarks Track (Round 2)}, 2021.

\bibitem{DBLP:journals/corr/RonnebergerFB15}
O.~Ronneberger, P.~Fischer, and T.~Brox, ``U-net: Convolutional networks for biomedical image segmentation,'' \emph{CoRR}, vol. abs/1505.04597, 2015.

\bibitem{song2020denoising}
J.~Song, C.~Meng, and S.~Ermon, ``Denoising diffusion implicit models,'' \emph{arXiv preprint arXiv:2010.02502}, 2020.

\bibitem{He2015DeepRL}
K.~He, X.~Zhang, S.~Ren, and J.~Sun, ``Deep residual learning for image recognition,'' \emph{2016 IEEE Conference on Computer Vision and Pattern Recognition (CVPR)}, pp. 770--778, 2015.

\bibitem{chi2023diffusion}
C.~Chi, Z.~Xu, S.~Feng, E.~Cousineau, Y.~Du, B.~Burchfiel, R.~Tedrake, and S.~Song, ``Diffusion policy: Visuomotor policy learning via action diffusion,'' \emph{The International Journal of Robotics Research}, p. 02783649241273668, 2023.

\bibitem{JMLR:v9:vandermaaten08a}
L.~van~der Maaten and G.~Hinton, ``Visualizing data using t-sne,'' \emph{Journal of Machine Learning Research}, vol.~9, no.~86, pp. 2579--2605, 2008.

\bibitem{Heusel2017GANsTB}
M.~Heusel, H.~Ramsauer, T.~Unterthiner, B.~Nessler, and S.~Hochreiter, ``Gans trained by a two time-scale update rule converge to a local nash equilibrium,'' in \emph{Neural Information Processing Systems}, 2017.

\bibitem{bińkowski2018demystifying}
M.~Bińkowski, D.~J. Sutherland, M.~Arbel, and A.~Gretton, ``Demystifying {MMD} {GAN}s,'' in \emph{International Conference on Learning Representations (ICLR)}, 2018.

\end{thebibliography}

\end{document}